\documentclass[journal]{IEEEtran}
\pdfoutput=1
\usepackage{cite}
\usepackage{mathptmx}

\usepackage{graphicx}
\usepackage{subfig}
\usepackage{caption}
\usepackage{epstopdf}
\usepackage{mathptmx} 
\usepackage{times} 
\usepackage{amsmath} 
\usepackage{amssymb}  
\usepackage{fancybox}
\usepackage{lineno}
\usepackage{amsthm}
\usepackage{algorithm}
\usepackage{algorithmic}
\usepackage{multirow}
\usepackage{longtable}
\usepackage{cleveref}
\usepackage{CJKutf8}
\usepackage{tablefootnote}
\AtBeginDocument{\begin{CJK}{UTF8}{gbsn}}
	\AtEndDocument{\end{CJK}}

\usepackage{booktabs}
\usepackage{threeparttable}
\allowdisplaybreaks

\newcommand{\bn}{\begin{eqnarray}}
\newcommand{\ed}{\end{eqnarray}}
\newcommand{\bnn}{\begin{eqnarray*}}
	\newcommand{\edd}{\end{eqnarray*}}
\newcommand{\by}{\begin{array}}
	\newcommand{\ey}{\end{array}}

\newcommand\old[1]{}

\ifCLASSINFOpdf
\else
\fi
\begin{document}
%
\title{RL-CSDia: Representation Learning of Computer Science Diagrams}
%
%
%

\author{Shaowei Wang, LingLing Zhang, Xuan Luo, Yi Yang, Xin Hu, and Jun Liu
}

\maketitle

\begin{abstract}
Recent studies on computer vision mainly focus on natural images that express real-world scenes. They achieve outstanding performance on diverse tasks such as visual question answering. Diagram is a special form of visual expression that frequently appears in the education field and is of great significance for learners to understand multimodal knowledge. Current research on diagrams preliminarily focuses on natural disciplines such as Biology and Geography, whose expressions are still similar to natural images. Another type of diagrams such as from Computer Science is composed of graphics containing complex topologies and relations, and research on this type of diagrams is still blank. The main challenges of graphic diagrams understanding are the rarity of data and the confusion of semantics, which are mainly reflected in the diversity of expressions. In this paper, we construct a novel dataset of graphic diagrams named Computer Science Diagrams (CSDia). It contains more than 1,200 diagrams and exhaustive annotations of objects and relations. Considering the visual noises caused by the various expressions in diagrams, we introduce the topology of diagrams to parse topological structure. After that, we propose Diagram Parsing Net (DPN) to represent the diagram from three branches: topology, visual feature, and text, and apply the model to the diagram classification task to evaluate the ability of diagrams understanding. The results show the effectiveness of the proposed DPN on diagrams understanding. 
\end{abstract}


%
\IEEEpeerreviewmaketitle

\section{Introduction}
\label{intorduction}

In recent years, some research on computer vision including image classification\cite{lu2007survey}, semantic segmentation\cite{long2015fully}, and visual question answering \cite{antol2015vqa} have been hot spots. Most of the studies are limited to natural images on datasets such as COCO\cite{lin2014microsoft}, Flickr\cite{young2014image}, and Visual Genome\cite{krishna2017visual} constructed from real-world scenes. However, driven by \textit{intelligent education}, novel tasks including multimodal knowledge fusion\cite{atrey2010multimodal}, textbook question answering\cite{kembhavi2017you} emerge in the computer vision community. Effectively capture the knowledge in diagrams and understand them are critical technologies of the above mentioned tasks.

\begin{figure}[t]
	\centering
	\vspace{0em}
	\includegraphics[width=0.9\linewidth]{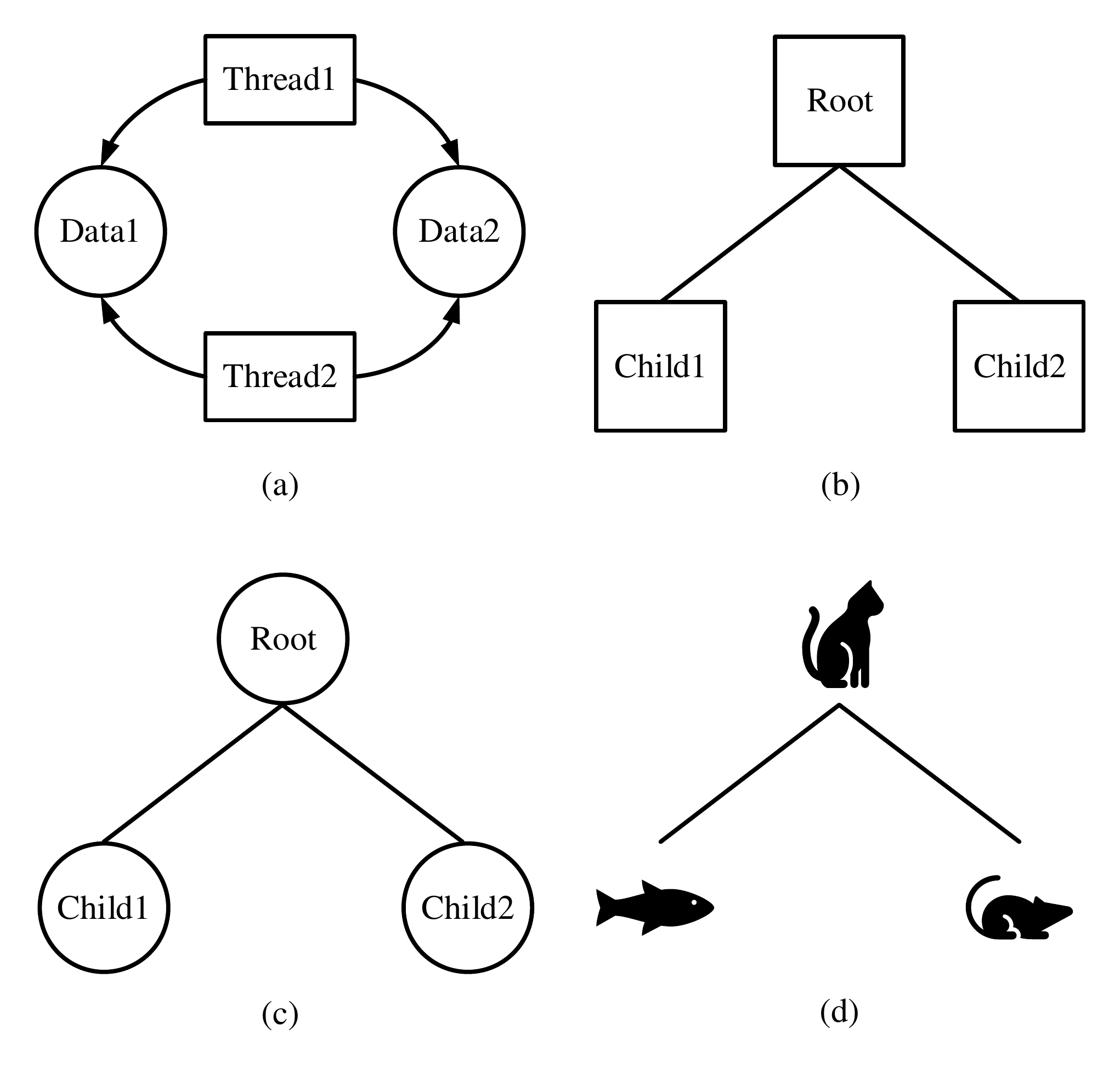}
	\caption{Diagrams examples of deadlock: (a) and binary tree: (b), (c), and (d).}
	\label{fig_amb}
\end{figure}

Diagrams are an extremely common visual form in the education field, they express various knowledge concepts in the educational scenes with more abstract semantics. They mostly exist in textbooks, blogs, and encyclopedias. Diagrams can be divided into two types according to their constituent elements. Existing research mainly focuses on the first type, which is from the Biology, Geography and other natural disciplines. The expression of these diagrams is similar to the natural images. Taking the food chain diagram as an example, the objects in it are mainly composed of things in natural scenes such as animals and plants. As shown in Fig. \ref{fig_amb}, the second type is composed of graphic objects, such as circles, rectangles, and triangles. These diagrams are visually simple, but contain rich semantic information. The relations between objects are no longer limited to spatial relations, but including complex logical relations. In Fig. \ref{fig_amb}a, the arrows between the threads and the data of deadlock indicate the required relations. In Fig. \ref{fig_amb}b, the root node and leaf nodes of the binary tree have parent-child topological relations. Understanding the second type of the diagrams is more challenging, mainly due to the following two challenges.

%

The first challenge is \textbf{semantic confusion}. The graphic itself does not have specific semantic, only when it is attached to a certain knowledge concept. This is mainly reflected in two problems: the first problem is similar to the polysemous words in natural language processing, which is one object can be given different semantics under different knowledge concepts. For example, the rectangle has different meanings in the deadlock and the binary tree. In Fig. \ref{fig_amb}a, it can represent the thread of the deadlock. While in Fig. \ref{fig_amb}b, the rectangle can represent the root node or leaf node of the binary tree. The second problem is similar to the synonymous words, which means one knowledge concept in diagrams has various expressions. When expressing a binary tree with the same topology in Computer Science domain, its nodes can be represented by rectangles (Fig. \ref{fig_amb}b), circles (Fig. \ref{fig_amb}c), or simple strokes (Fig. \ref{fig_amb}d). 

The second challenge is lack of data. High-quality data that can be used in the research on diagram is difficult to obtain, relevant datasets are very scarce. The main reason is that the annotation of the diagrams is complicated. Firstly, due to the diagram involves a wide range of subjects and contains complex knowledge concepts, the annotators need rich knowledge backgrounds, and it takes long time to read and understand the content of the corresponding textbooks. In order to make the annotation form and content consistent, it is necessary to formulate terminology rules for incompatible subjects. Secondly, diagrams annotation needs a finer granularity than the natural images because of the semantic confusion. Using the annotation method in this paper, it takes an average of twenty minutes for a graduate student to annotate a diagram in Computer Science domain, while the annotation of natural images is often instantaneous.

In this paper, we construct a dataset named Computer Science Diagrams (CSDia). It contains 1,294 diagrams from Computer Science courses, with more than 30,000 rich annotations. CSDia is the first diagram dataset in Computer Science domain, which can be used in different tasks such as object detection and multimodal knowledge fusion. Compared with other existing diagram datasets, CSDia is composed of geometric shapes, which are more challenging to understand due to semantic confusion and lack of data. Hence we propose a novel method for generating the topology of diagram, and make a preliminary attempt to understand these diagrams. This method reduces the visual noise caused by variety of expressions, and enhances the topological structures of diagrams. In the condition, we apply the topology into a new classification model Diagram Parsing Net (DPN) that also combines visual feature and text, to test whether the method understands the diagrams well. 

Our contributions include: (a) A new diagram dataset named CSDia for visual research in the Computer Science domain; (b) A novel method of generating the topology to parse the diagrams; (c) A model for diagram classification which considers CNN features, topology, and text information of the diagrams in three branches, and is evaluated with baselines on the CSDia dataset.

\section{Related work}
\label{Related work}

Diagrams are widely used as a form of expression in educational resources to intuitively express the characteristics of the knowledge concepts in visual form, and make learners better understand the connotation of the concepts. Diagrams understanding is of great significance, but little attention is paid to it. 

Specifically, the research of diagrams originated in the 1990s. In the early days, researchers generally used traditional rule-based methods to study diagrams. They completed tasks such as diagram analysis, classification, and geometric question answering. Watanabe et al.\cite{watanabe1998diagram} proposed a method for analyzing the pictorial book of flora (PBF) diagrams by using natural language information and layout information. The limitation is the inefficient way of using handwritten rules to extract and represent the diagrams. Ferguson et al.\cite{ferguson2000georep} created a spatial reasoning engine to generate qualitative spatial descriptions from line drawings. They also proposed a model of repetition and symmetry detection which can model human cognitive process when reading repetition based diagrams\cite{ferguson1998telling}. Later, Futrelle et al.\cite{futrelle2003extraction} studied the extraction of the diagrams from PDF documents, and performed a classification task on it, but only for bar, non-bar diagrams. As for geometric question answering problems, Seo et al.\cite{seo2014diagram} identified visual elements in a diagram while maximizing agreement between textual and visual data to build an automated system that can solve geometry questions. Sachan et al.\cite{sachan2017learning} used detailed demonstrative solutions in natural language to solve geometry problems using axiomatic knowledge.

Recent years, methods based on deep learning have been widely used in diagram studies, such as textbook question answering and illustrations classification tasks. Specifically, Kembhavi et al.\cite{kembhavi2016diagram} introduced the Diagram Parse Graphs (DPG) as the representation to model the structure of diagrams and used it for semantic interpretation and reasoning tasks. The experiments were conducted on AI2 Diagrams (AI2D) dataset which contains diagrams from elementary school science textbooks. They also tested three different types of deep learning models on the textbook question answering task on Textbook Question Answering (TQA) dataset containing diagrams from life, earth and physics textbooks. So far, the AI2D and TQA datasets are the most widely used for diagram-related tasks\cite{gomez2019look,gomez2020isaaq,kim2018textbook}. Later, Morris et al.\cite{morris2020slideimages} used a standard deep neural architecture MobileNetV2\cite{sandler2018mobilenetv2} achieve the task of classifying educational illustrations on the dataset named SlideImages.

To sum up, early rule-based methods are often used for specific data such as histograms and non-histograms. The methods are inefficient and have limited capabilities for diagrams representation. Deep learning-based methods solve more difficult tasks, but the datasets used for verification such as AI2D and TQA still focus on natural disciplines which are quite different from the diagrams we try to explore. Therefore, it is of great significance to construct a diagram dataset composed of pure geometric shapes and study how to parse them.
 
 \begin{table}[t]
 	\caption{Sources of CSDia dataset.\label{sources}}
 	\centering
 	\center
 	\renewcommand{\arraystretch}{1.4}
 	\tabcolsep=2.0 pt
 	\begin{tabular}{c|c}
 		\toprule
 		\multirow{1}{*}{English} & \textit{Data Structure and Algorithm Analysis in C}\cite{shaffer2012data}          \\
 		\multirow{1}{*}{textbook}& \textit{Algorithms and Data Structures: The Basic Toolbox}\cite{mehlhorn2008algorithms}     \\ \hline
 		& \textit{数据结构高分笔记 (High Score Notes of Data Structure)}\cite{shuaihui2018}                 \\
 		\multirow{1}{*}{Chinese} & \textit{数据结构C语言版 (Data Structure C version)}\cite{yanweimin2002}                  \\
 		\multirow{1}{*}{textbook}& \textit{计算机操作系统 (Computer Operating System)}\cite{tangxiaodan2007caozuoxitong}            \\
 		& \textit{计算机组成原理 (Principles of Computer Organization)}\cite{tangshuofei2000zucheng} 					\\
 		& \textit{数字逻辑电路 (Digital Logic Circuit)}\cite{liuchangshu2002shuzi}                                \\ \hline
 		\multirow{3}{*}{Blog}    & Zhihu\tablefootnote {https://www.zhihu.com}                                                         \\
 		& Chinese Software Developer Network\tablefootnote {https://www.csdn.net}                              \\
 		& Douban\tablefootnote {https://www.douban.com}                                                        \\ \hline
 		\multirow{2}{*}{Encyclopedia}     & Baidu pedia\tablefootnote {https://baike.baidu.com}                                                    \\
 		& Wiki pedia\tablefootnote {https://www.wikipedia.org}                                                    \\
 		\bottomrule
 	\end{tabular}
 	\vspace{-1em}
 \end{table}
 
 \begin{figure*}[t]
 	\centering
 	\vspace{-2em}
 	\includegraphics[width=1\linewidth]{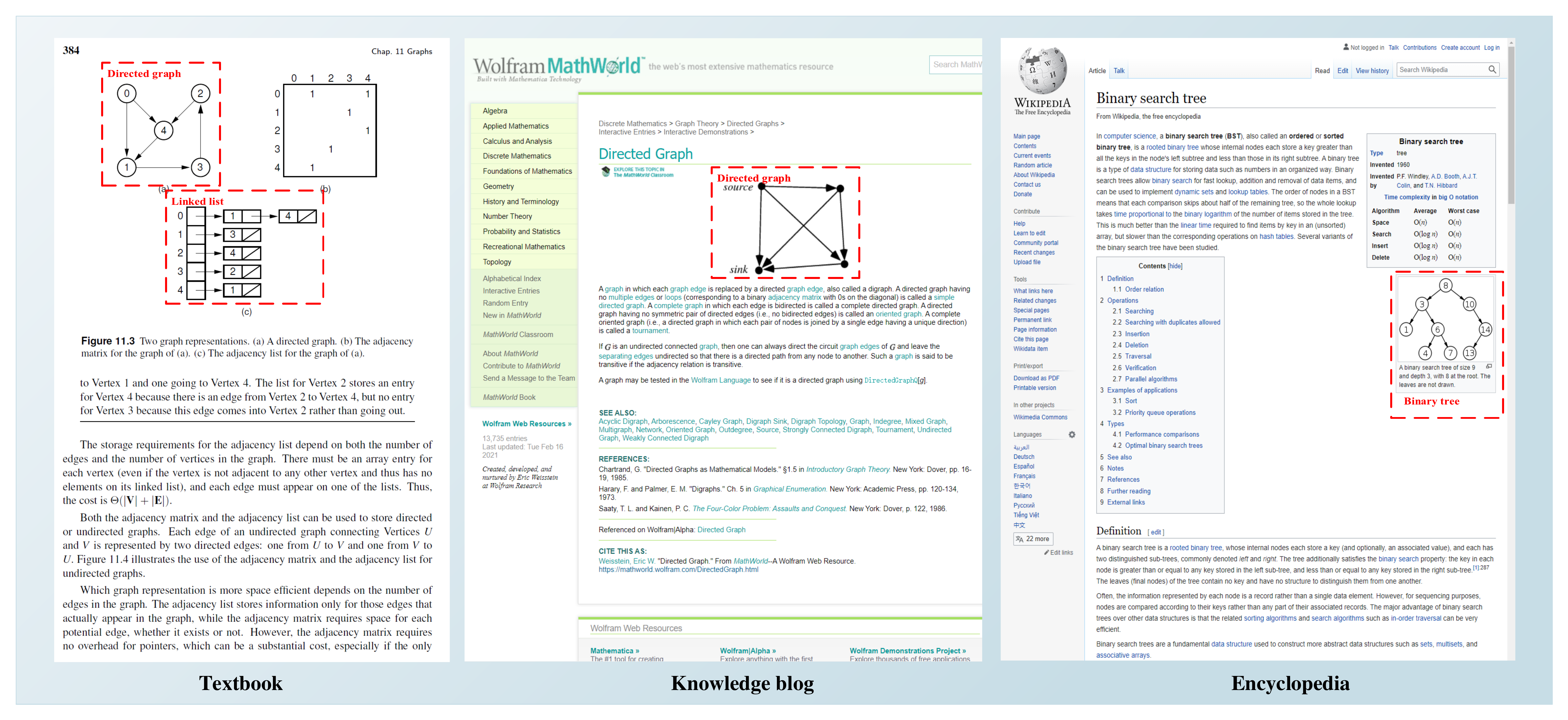}
 	\caption{Examples of various sources in CSDia.}
 	\label{fig_sources}
 \end{figure*}

\begin{figure*}[t]
	\centering
	\vspace{-2em}
	\includegraphics[width=1\linewidth]{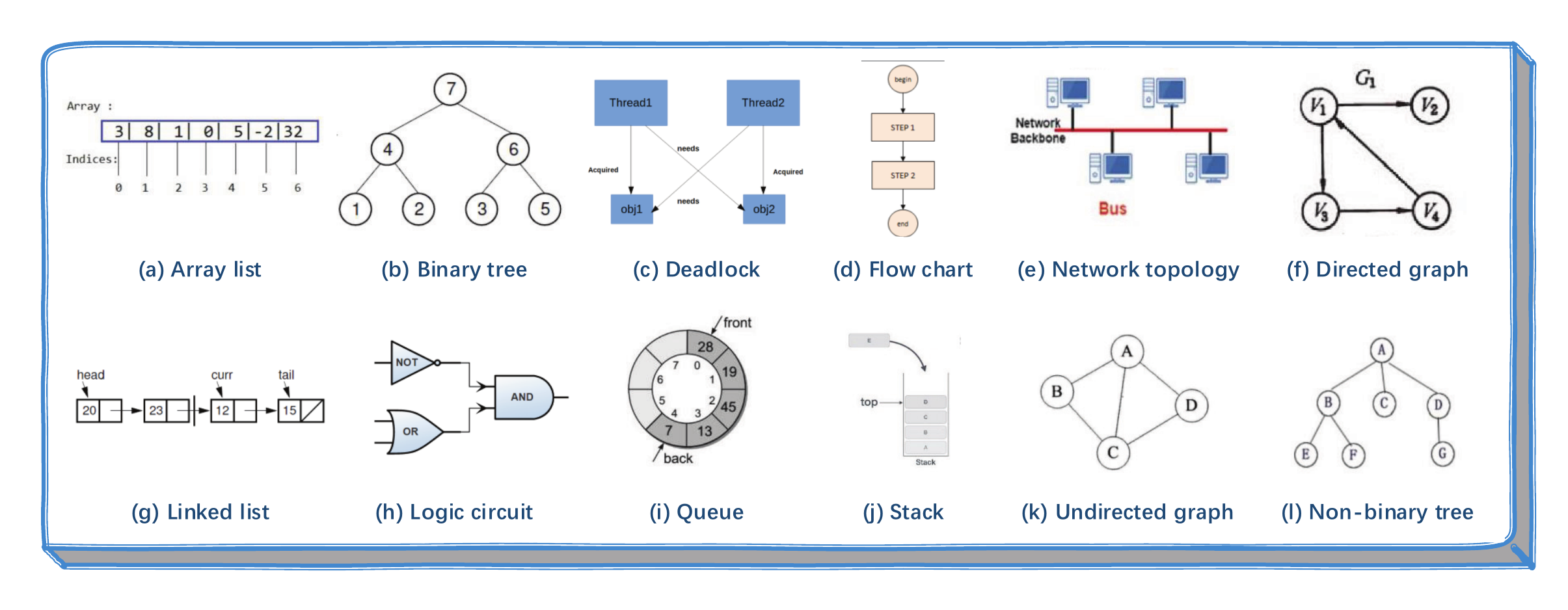}
	\caption{Examples of categories in CSDia.}
	\label{fig_csdia}
\end{figure*}

\section{The CSDia Dataset}
\label{CSdia dataset}

In this section, we introduce the construction process of CSDia dataset. The process contains two procedures: (1) Diagram collection; (2) Diagram annotation, which contains global attributes, objects and relations. We introduce the above two procedures in detail in \ref{diagram collection} and \ref{annotation}. The construction of this dataset takes five months in total. We recruited twelve students as annotators from the department of Computer Science, including four third grade undergraduates, two fourth grade undergraduates, and six graduate students. We confirm these annotators have taken relevant courses of the diagrams and have qualified knowledge background. The annotators use unified concepts according to the textbook \textit{Data Structure and Algorithm Analysis} as the first benchmark.

\subsection{Diagram Collection}
\label{diagram collection}

Due to the scarcity of diagrams, we use a multi-source method to collect them. Specifically, we select textbooks, blogs, encyclopedias as data sources of the CSDia other than crawling. See TABLE \ref{sources} for detailed sources. Examples of various sources are shown in Fig. \ref{fig_sources}.

\begin{figure*}[t]
	\centering
	\vspace{-2em}
	\includegraphics[width=1\linewidth]{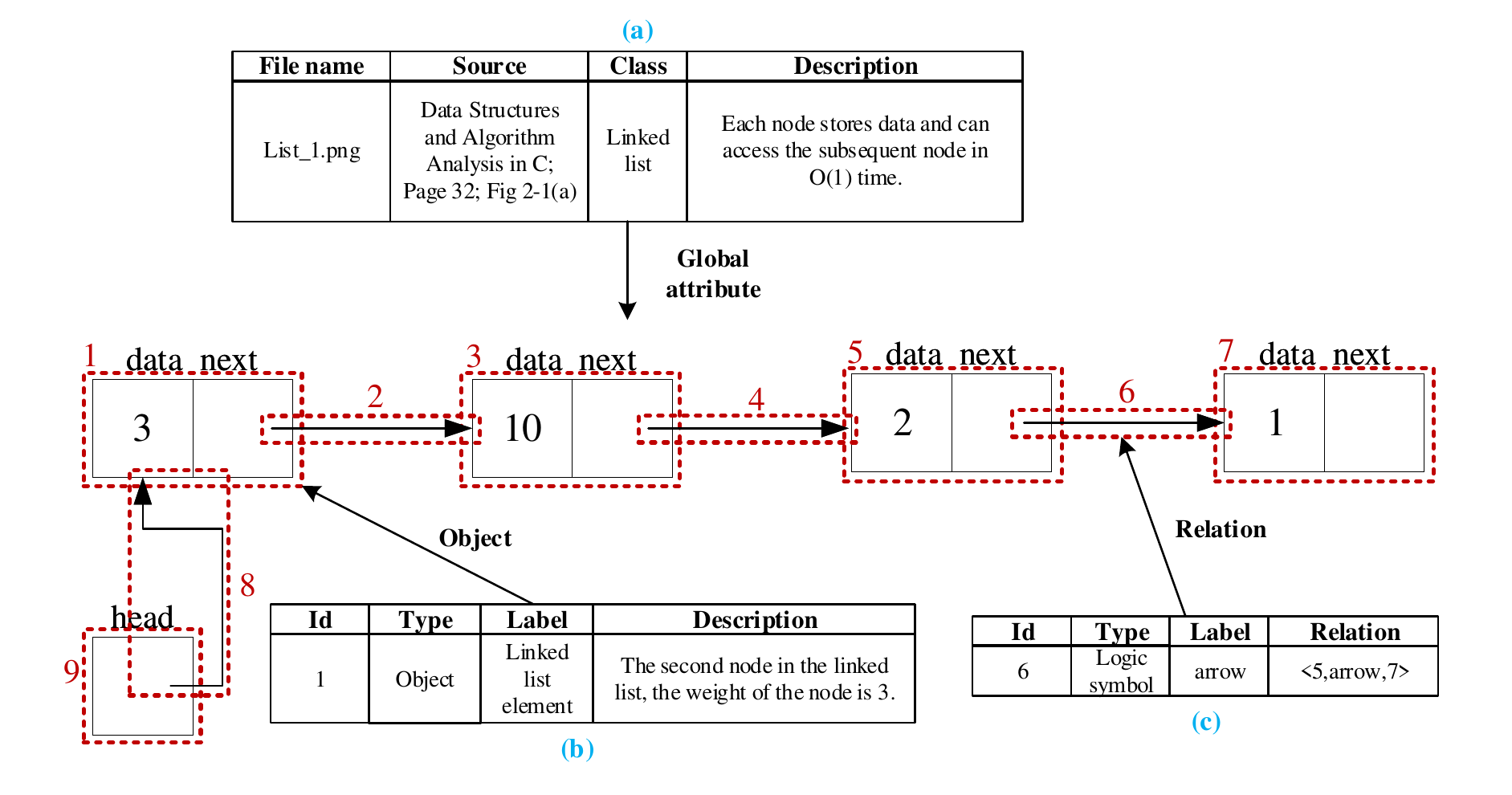}
	\caption{Example of a linked list diagram annotation in CSDia.}
	\label{fig_ann}
\end{figure*}

In order to ensure the quality of the collected diagrams, we adopt a semi-automatic filtering method to purify the preliminary collected data. Specifically, for high-quality diagrams sources such as textbooks, encyclopedias, and blogs, the diagrams are directly manually selected by the annotators. We use these data as positive examples of diagrams, and various scenes in the COCO dataset as negative examples of natural images, to train a binary classifier (diagrams-natural images classifier). We use VGG16 \cite{simonyan2014very} as the basic network, without using the pre-trained model. The images are processed with gray scale as inputs, and finally the accuracy of filtering diagrams can reach 99.46\%. We use the trained model for the data obtained by the search engine crawling. Finally, the data from all sources can reach a higher quality. Examples of each category in the CSDia dataset are shown in the Fig. \ref{fig_csdia}.


\subsection{Diagram Annotation}
\label{annotation}

Consistent with the cognitive law of things, we set the annotation process of the diagram from global to local, which also makes the annotation information to assist tasks at different granularities. As shown in Fig. \ref{fig_ann}, drawing on the way of natural images annotation, we set set fine-grained annotations on the global attributes, objects and relations. 

\textbf{Global attribute.} The premise of the vast majority of diagram-related research is to have an overall understanding of it. The content summarizes the knowledge concepts and describes the information of the diagrams macroscopically. This part of the annotation includes source, class and description. As Fig. \ref{fig_ann}a shows, the \textit{source} records current collection of the diagram in detail. When coming from a textbook, the book title, page number are labeled. If it comes from the Internet, the URL is labeled. The \textit{class} indicates the knowledge unit to which the diagram belongs. The \textit{description} is a brief textual summary of the knowledge unit expressed by the diagram. If the description is in the source of the diagram, it is recorded directly, such as the contextual explanation in the textbook. If not, it is summarized manually based on relevant knowledge.

\textbf{Object.}
The objects in a diagram refer to the constituent elements in the current knowledge concept, such as the nodes of the binary tree and the head of the queue. Most of the objects in the diagrams of Computer Science are geometric shapes, and some text information are attached as supplement descriptions around them. As shown in Fig. \ref{fig_ann}b, we use rectangular bounding boxes to locate the objects in a four-dimensional coordinate form. Then we give each object four-aspect annotations, including \textit{id}, \textit{type}, \textit{label}, and \textit{description}. Among them, the \textit{id} is a unique identification number starting from one for each object. The \textit{type} means that we divide objects into two groups: semantic shapes (such as circular stack node) and logical symbols (such as arrows). The \textit{label} is the subcategory of the object under the knowledge unit. The \textit{description} contains other related information of the objects, such as the weight of the node, the head node of the queue, or the text around the object.

\textbf{Relation.}
Diagrams depict complex phenomena and higher-order relations between objects that go well beyond what a single natural image can convey\cite{kembhavi2016diagram}, such as link relations between linked list nodes, and parent-child relations between binary tree nodes. Due to this characteristic, we attach annotations to various logical relation symbols, such as arrows, lines. Similarly, we first use rectangular bounding boxes to locate the logical symbols. We give each relation three-aspect annotations, including \textit{id}, \textit{label} and \textit{relation triple}. Among them, the \textit{id} is a unique identification number for each relation. The \textit{label} indicates the type of symbols on which the relation depends, such as arrow and line. The \textit{relation triple} indicates the objects and the logic symbol involved in a relation, such as $<$1,arrow,3$>$ in Fig. \ref{fig_ann}c.

\subsection{Statistic}
\label{Statistic}

\begin{table}[t]
	\caption{Detailed statistics for each category in CSDia.}
	\centering
	\center
	\renewcommand{\arraystretch}{1.4}
	\tabcolsep=13.0 pt
	\begin{tabular}{cccccccc}
		\cline{1-4}
		Category       		& Diagrams     & Objects    & Relations     &  \\ \cline{1-4}
		Array list  		& 100          & 583  		& 468         	&  \\
		Linked list     	& 74           & 626     	& 375         	&  \\
		Binary tree    		& 150          & 1,323    	& 590 			&  \\
		Non-Binary tree 	& 150		   & 1,489 		& 651          	&  \\ 
		Queue    			& 150          & 1,261    	& 444 			&  \\
		Stack    			& 150          & 540    	& 403 			&  \\
		Directed graph    	& 71           & 695    	& 377 			&  \\
		Undirected graph   	& 79           & 828    	& 437 			&  \\
		Deadlock    		& 100          & 840    	& 423 			&  \\
		Flow chart    		& 100          & 985   		& 458 			&  \\
		Logic circuit    	& 70           & 913  		& 432 			&  \\
		Network topology   	& 100          & 1,593   	& 517 			&  \\\cline{1-4}
		Total    				& 1,294         & 11,776   	& 5,675 			&  \\\cline{1-4}	\label{sta}
	\end{tabular}%
\vspace{-2em}
\end{table}

CSDia dataset contains a total of 1,294 diagrams in 12 categories from five undergraduate courses: \textit{Data structure}, \textit{Principles of Computer Networks}, \textit{Computer Architecture}, \textit{Digital Logic Circuit}, and \textit{Computer Operating System}. On the whole, CSDia contains annotations of more than 11,000 objects and 5,600 relations. As shown in TABLE \ref{sta}, each category of the dataset is unbalanced, ranging from 71 to 150 per class. We split diagrams into 951 for training sets and 343 for test sets.

\section{Approach}
\label{approach}

In this section, we propose a new model for the diagram classification task. The model combines the CNN features of the diagram, CNN features of the topology and text information from three branches. It improves the ability of traditional deep models extracting features from diagrams with abstract representations. In our model, the topology branch is the key point. This branch can extract effective features such as objects positions and structural relations, and provide support for understanding the diagram. We introduce the framework and logic topology in \ref{diagram parsing net} and \ref{generation of the logic structure} respectively.

\subsection{Diagram Parsing Net}
\label{diagram parsing net}

\begin{figure}[t]
	\centering
	\vspace{0em}
	\includegraphics[width=0.98\linewidth]{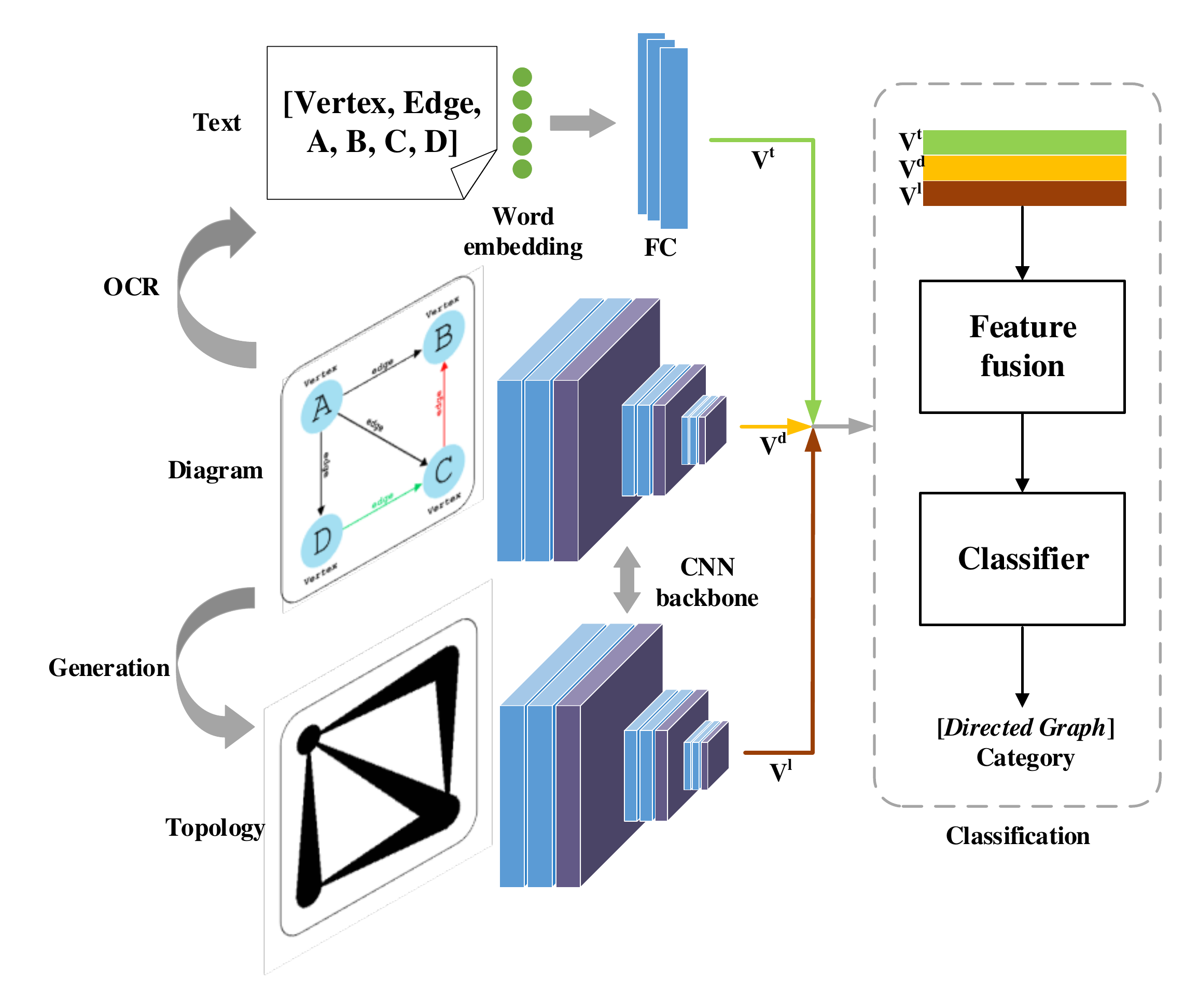}
	\caption{Overview of the DPN model for diagram classification.}
	\label{fig_model}
\end{figure}

As shown in Fig. \ref{fig_model}, we propose DPN to achieve the classification task on CSDia. The model is divided into three branches to analyze the diagrams. With a input diagram, we consider its RGB feature $X$. Firstly, in the original diagram branch of model (central of Fig. \ref{fig_model}), we use the traditional CNN network such as ResNet for feature extraction:
\begin{align}
	\mathbf{v}^d = f_{\theta}(X),
\end{align}
where $f_\theta(\cdot)$ is the deep non-linear function with the parameter set $\theta$ for the CNN network, $\mathbf{v}^d$ is the embedding vector of the input diagram. 

Secondly, because most of the diagrams are accompanied by relevant text information, which plays a great role in the understanding of the knowledge unit, we use text branch (up part of the Fig. \ref{fig_model}) to parse it. With Optical Character Recognition (OCR) technology, the words identified from the diagram are in the set $T = \{t_i | i=1,...,k\}$. We use the pre-trained GloVe \cite{pennington2014glove} model to convert the extracted words into vectors:
\begin{align}
	\mathbf{w}_i = f_{GloVe}(t_i),i=1,...,k,
\end{align}
\begin{align}
	\mathbf{x}^t = Pool([\mathbf{w}_{1};...;\mathbf{w}_{k}]),
\end{align}
where $f_{GloVe}(\cdot)$ is the function of pre-trained GloVe model,  $[\cdot]$ is the concatenation operation, $\mathbf{w}_i$ is the vector of each word $t_i$ after embedding, $\mathbf{x}^t$ is the vector after average pooling function of all word vectors in the diagram. We use a fully connected layer to process the resulting vector $\mathbf{x}^t$.
\begin{align}
	\mathbf{v}^t = Relu(W_t\mathbf{x}^t+\mathbf{b}_t),
\end{align}
where $W_t$ is the weight matrix to be optimized, $\mathbf{b}_t$ is the bias vector, and $Relu(\cdot)$ is the activation function that limits the output element to be positive. In this condition, the representation vector of the text in the diagram is obtained as $\mathbf{v}^t$.

Thirdly is the topology branch in the down part of Fig. \ref{fig_model}, whose function is to extract the topological structures and objects locations information inside the diagram, thereby reducing the visual noises.
\begin{align}
	X^* = g(X,B), \label{generation}
\end{align}
where $g(\cdot)$ is the generation function of the topology with input $X$ and bounding boxes location of each object $B$. $X^*$ is the generated single channel diagram of the topology. Similarly, we use the same CNN network to process the topology:
\begin{align}
	\mathbf{v}^l = f_{\psi}(X^*),
\end{align}
where $f_\psi(\cdot)$ is the deep non-linear function with the parameter set $\psi$ for the CNN network, $\mathbf{v}^l$ is the embedding vector of the topology. 

Finally, the diagram $X$ is represented as the vector $\mathbf{r}$ that combines the three-branch information as follows:
\begin{align}
	\mathbf{r} = Relu(W_d[\mathbf{v}^d;\mathbf{v}^t;\mathbf{v}^l]+\mathbf{b}_d),
\end{align}
where $[\cdot]$ is the concatenation operation, $W_d$ is the weight matrix to be optimized, and $\mathbf{b}_d$ is the bias vector. Then we feed the vector into the classifier to get the category probability distribution vector $\mathbf{s}$:
\begin{align}
	\mathbf{s} = Softmax\left(f_\phi(\mathbf{r})\right),
\end{align}
where $f_{\phi}(\cdot)$ is the multi-layer perceptron (MLP) network that takes one vector as input and includes the same hidden neurons as the categories at the output layer with the parameters $\phi$. The function $Softmax(\cdot)$ is to normalize the output variable of MLP for probability of the category. Finally, the category corresponding to the maximum value in $\mathbf{s}$ is the result of classification.

\subsection{Generation of the Topology}
\label{generation of the logic structure}

\begin{figure}[t]
	\centering
	\vspace{0em}
	\subfloat[Directed graph and its topology]{
		\includegraphics[width=0.4\linewidth]{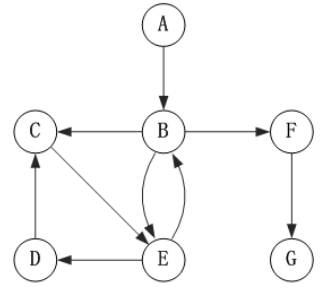}\quad
		\includegraphics[width=0.4\linewidth]{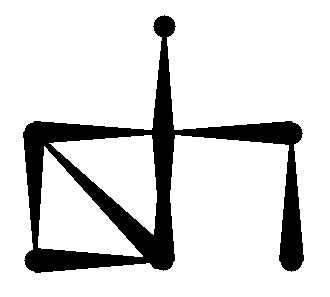}}

	\subfloat[Undirected graph and its topology]{
		\includegraphics[width=0.4\linewidth]{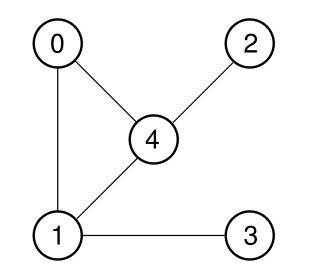}\quad
		\includegraphics[width=0.4\linewidth]{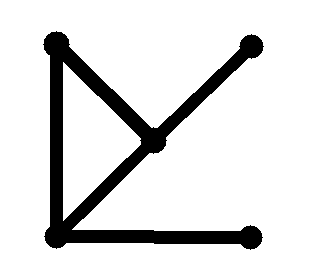}}
	\caption{Examples of logic structure generation methods for directed edges and undirected edges. \label{fig:logic_structure}}
\end{figure}

The topology is a highly abstract and general expression of the diagram, which is the $g(\cdot)$ in Equation (\ref{generation}). It removes the noise caused by different visual elements, and only retains the locations and structural relations in the diagrams. For example, three binary trees in Fig. \ref{fig_amb} express the same topology but have various forms of visual expressions. When classifying the knowledge concepts represented by these diagrams, the information brought by different visual expressions is redundant. Therefore, the diagram need to be generalized to a more certain extent, that is, the topology.

In essence, the topology is to reconstruct a diagram in a unified form after extracting the objects and relations of the original diagram. Among them, we use circles to represent all objects as abstract objects. Firstly, all objects in a diagram can be expressed as a set $O = \{o_i | i=1,...,k\}$, the central coordinate of the generated circle $(x_i,y_i)$ according to object $o_i$ is determined as the following way:
\begin{equation}
x_i = (W^r_i - W^l_i) / 2,
\end{equation}
\begin{equation}
y_i = (H^u_i - H^l_i) / 2,
\end{equation}
where $W^r_i$ and $W^l_i$ are the coordinates of the right and left borders of the object's bounding box respectively, $H^u_i$ and $H^l_i$ are the coordinates of the upper and lower borders of the object's bounding respectively. Then, the radius of the circle $r_i$ is defined in the following method:
\begin{equation}
B^W_i = W^r_i - W^l_i,
\end{equation}
\begin{equation}
B^H_i = H^u_i - H^l_i,
\end{equation}
\begin{equation}
r_i = \lambda^2_{r} * \sqrt{B^H_i * B^W_i}/\pi \label{radius},
\end{equation}
where $B^W_i$ and $B^H_i$ are the width and height of bounding box, respectively. In Equation (\ref{radius}), $\lambda_{r}$ is a regularization parameter. The purpose is to normalize the size of the generated circle with respect to the entire canvas, so as to avoid the situation of being too large or too small. It can be calculated by the following formula:
\begin{equation}
\lambda_{r} = \sqrt[8]{H * W} / 10,
\end{equation}
where $H$ and $W$ are the height and width of the entire canvas, respectively. The calculation formula of the normalization coefficient is an empirical formula. We find that when the times of root is larger, the content of the canvas is better distributed, so the method of eighth root is selected. In practical applications, it is appropriate to select the times greater than five.

Next step is to generate a representation of the relation between objects. In the diagrams, part of the relations are directed, such as a directed graph in data structure or a flow relation in a flowchart. Part of the relations are undirected, such as the edges in a binary tree. Whether the relation is directional or undirected, it is of great significance to the topology of the entire diagram. We consider the following method to distinguish them. First of all, for any relation, it is dependent on two objects. We call these two objects the head and the tail of the relation, which are the two circles generated by above method. The way to generate the abstract relation is to determine an edge with a certain width between the head object and the tail object. We generated undirected relations according to the following formula:
\begin{equation}
Line^u_{head} = Line^u_{tail} = (r_{head} + r_{tail})/2,
\end{equation}
Where $r_{head}$ and $r_{tail}$ are the radius of the circle of the head object and the tail object respectively. $Line^u_{head}$ and $Line^u_{tail}$ are the width of the relation line at the head and tail object, respectively. For directed relations, the calculation is as follows:
\begin{equation}
Line^d_{head} = 0,\ Line^d_{tail} = r_{tail},
\end{equation}
where $r_{tail}$ is the radius of the circle of the tail object. $Line^d_{head}$ and $Line^d_{tail}$ are the width of the relation line at the head and tail object, respectively. The topology generated by the above method is shown in Fig. \ref{fig:logic_structure}. It can be seen that the above method can well represent the structure and relation information of the diagram, and can distinguish different types of relations.

\section{Experiment}
\label{experiment}

In this section, we conduct rich experiments on the proposed CSDia dataset. \ref{Setting} is the experimental setting. \ref{Comparative} is a comparative analysis of classification methods. In \ref{Abalation}, we conduct ablation experiments to further analyze the contribution of each branch to the classification performance. In \ref{impact of ls}, we conduct dimensional and directional analysis of the topology branch.

\subsection{Experimental setting}
\label{Setting}

In all experiments, we use CNN models with parameters pretrained on ImageNet \cite{deng2009imagenet} and modify the output dimension of the last fully connected layer. When using OCR to extract text, we select a third-party library called EasyOCR \cite{easyocr}. The obtained text is embedded using the pre-trained GloVe model, and the embedding size is set to 50. A two-layer multi-layer perceptron (MLP) is used to reduce the dimensions of three branches with hidden layer size 80. We reduce the diagram feature dimension to 120, the logic structure feature dimension to 100, and the text feature to 40 in \ref{Comparative} and \ref{Abalation}. We uniformly use the SGD optimizer, with the learning rate 4e-3 for the first 30 epochs, 1e-4 for the last 30 epochs and momentum 0.9. All results are obtained by running after 20 times. The dataset follows the same split as mentioned in \ref{Statistic}. 

\subsection{Comparative Analysis}
\label{Comparative}

Due to the scarcity of datasets and differences in visual features, the research on diagram classification is still blank. Therefore, in the comparative analysis, we select four state-of-the-art models in the classification of natural images.
\begin{itemize}
	\item \textbf{ResNet} \cite{he2016deep}: The core idea is the residual connection that skips one or more layers. The motivation for skipping over layers is to avoid the problem of vanishing gradients. Because of its compelling results, ResNet becomes one of the most popular architectures in various computer vision tasks. We use the 50-layer version of the ResNet for all the experiments in this paper.
	\vspace{1.0ex}
	
	\item \textbf{ResNeXt} \cite{xie2017aggregated}: It is a variant of ResNet, a simple, highly modularized network architecture for image classification. The network is constructed by repeating a building block that aggregates a set of transformations with the same topology.
	\vspace{1.0ex}
	
	\item \textbf{SqueezeNet} \cite{iandola2016squeezenet}: This is a lightweight and efficient CNN model for image classification. It has 50 times fewer parameters than AlexNet and maintains AlexNet-level accuracy on ImageNet without compression.
	\vspace{1.0ex}	
	
	\item \textbf{MobileNetV2} \cite{sandler2018mobilenetv2}: It is an improved version of MobileNet, which uses linear bottlenecks and inverted residuals technology to further improve the performance.
	\vspace{1.0ex}
\end{itemize}

\begin{table}[t]
	\caption{Comparison results.}
	\centering
	\center
	\renewcommand{\arraystretch}{1.4}
	\tabcolsep=10.0 pt
	\begin{tabular}{cccccccc}
		\cline{1-4}
		Model       & Acc              & Our model            & Acc              &  \\ \cline{1-4}
		SqueezeNet  & 53.81          & DPN with SqueezeNet  & 81.25          &  \\
		ResNext     & 76.82          & DPN with ResNext     & 89.93          &  \\
		ResNet50    & 77.60          & DPN with ResNet50    & 92.36 &  \\
		MobileNetV2 & 83.16 & DPN with MobileNetV2 & 89.14          &  \\ \cline{1-4}
	\end{tabular} \label{performance}
\end{table}

TABLE \ref{performance} shows the classification performance of the four CNN models used independently and as a backbone in the DPN model. The results show that no matter which CNN model is used, the performance of the DPN has improved when compared with the traditional natural image classification models. On the one hand, when using MobileNetV2,  the accuracy of the DPN model increases the least by 6.25\%, and it increases the most by 27.44\% when using SqueezeNet. On the other hand, DPN gets its best classification performance 92.36\% when using ResNet as backbone. The reason why DPN has a more obvious performance improvement is it considers the topological information, text information related to the diagram, which play important roles in the understanding of the diagrams.

\subsection{Abalation Study}
\label{Abalation}
\begin{table}[t]
	\caption{Results different combinations of the modules in DPN for classification.}
	\centering
	\center
	\renewcommand{\arraystretch}{1.4}
	\tabcolsep=25.0 pt
	\begin{tabular}{cccccccc}
		\cline{1-2}
		Module                                       & Acc                    &  \\ \cline{1-2}
		Original diagram                             & 77.60                &  \\
		Logic structure                              & 78.82	              &  \\
		Original diagram + Logic structure           & 89.11                &  \\
		Original diagram + Text                      & 83.85                &  \\ 
		Logic structure + Text                       & 86.54                &  \\ 
		Original diagram + Logic structure + Text    & \textbf{92.36}       &  \\ \cline{1-2}
	\end{tabular} \label{tabel_abalation}
\end{table}

In the DPN model, we analyze the diagram from three branches, namely the original diagram, the topology of diagram, and the text in the diagram. In order to explore the contribution of each branch, we design six model variants as shown in TABLE \ref{tabel_abalation}. By analyzing the experimental results, we can draw the following conclusions:

\begin{itemize}
	\item It is difficult to fully understand the diagram by only extracting the CNN features due to its confusion of visual features. It only has a classification accuracy of 77.60\% when only the CNN features of diagram is input, which is about 15\% lower than using the complete DPN model.
	\vspace{1.0ex}
	
	\item The topology of diagram plays a significant role in understanding the diagram. When only the diagram is input for comparison, the classification accuracy improves 11.51\% when diagram and logic structure are input simultaneously. Even if only the topology is input, the accuracy is increased by 0.87\%. It indicates that the method of generating the topology effectively filters the noise at the level of visual elements, while retaining the effective information in the diagram.
	\vspace{1.0ex}
	
	\item Text information also plays a certain role in diagram understanding task. Compared with the diagram input only, the accuracy improves 6.25\% when diagram and text are input simultaneously. The reason is that the knowledge concept contained in the diagram cannot be conveyed only through pixels, but the text is also helpful.
	\vspace{1.0ex}	
\end{itemize}

\subsection{Impact of Topology}
\label{impact of ls}

\begin{figure}[t]
	\centering
	\vspace{-2em}
	\includegraphics[width=0.98\linewidth]{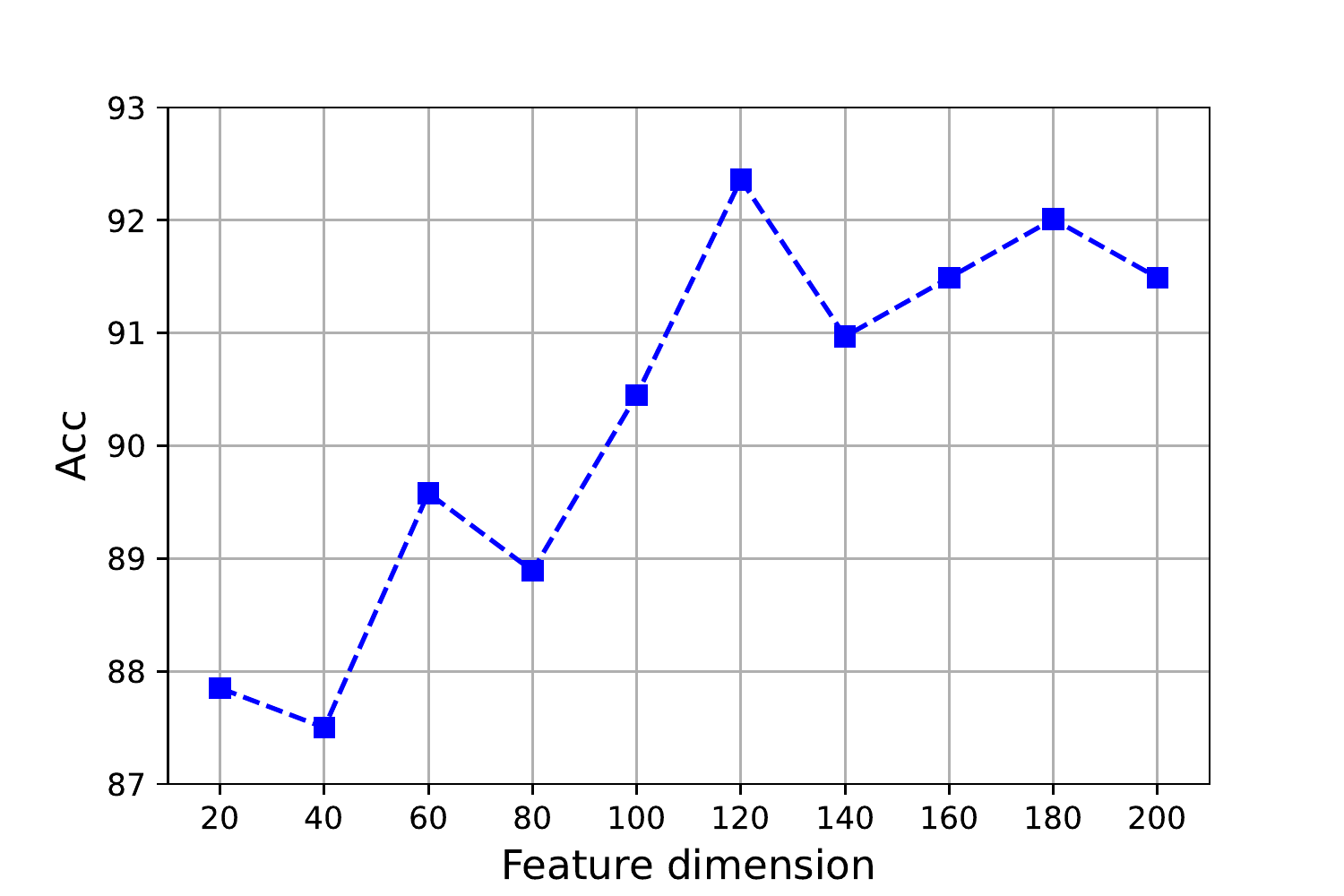}
	\caption{Analysis on the dimension of logic structure.}
	\label{fig_dimension}
\end{figure}

\begin{table}[t]
	\caption{Performance of directed and undirected logic structure.}
	\centering
	\center
	\renewcommand{\arraystretch}{1.4}
	\tabcolsep=6.0 pt
	\begin{tabular}{cccccccc}
		\cline{1-4}
	Type									&Category       		& Acc (Directed)     	& Acc (Undirected)     		&  \\\cline{1-4}
	\multirow{6}{*}{Directed}	&Deadlock    		& 57.14\%          		& 35.71    				&  \\
										&Directed graph	 	& 72.73		   		& 54.55 					&  \\ 
										&Flow chart    		& 100.00          	& 80.00    				&  \\
										&Linked list  		& 100.00          	& 88.89    				&  \\
										&Logic circuit  		& 100.00          	& 90.91    			&  \\
										&Network topology    & 78.75           	& 64.29    				&  \\\cline{1-4}
	\multirow{6}{*}{Undirected }		&Array list  		& 53.33          		& 60.00  					&  \\
										&Binary tree     	& 100          		& 100.00     				&  \\
										&Undirected graph    & 72.73           	& 81.82  					&  \\
										&Non-binary tree   	& 95.24           	& 100.00    				&  \\
										&Queue    			& 43.48          		& 47.83    				&  \\
										&Stack    			& 91.30          		& 69.57   				&  \\\cline{1-4}
										&All    				& 78.82         		& 71.88   				&  \\\cline{1-4}	\label{acc per cato}
	\end{tabular}%
	\vspace{-4em}
\end{table}

In DPN, the topology plays an extremely important role. In this section, we study the topology in detail from two aspects: dimension and direction. 

Firstly, while keeping the dimensions of the other two branches unchanged, the dimension of the topology $\mathbf{v}^l$ is varied in [20, 200] with the step of 20. The classification performance curves on CSDia are shown in Fig. \ref{fig_dimension}. The performance changes show a trend of increasing firstly and fluctuating later. The model has better performance when the dimension is greater than 100. The result indicates that it's suitable to set $\mathbf{v}^l$ over 100 when DPN is applied in practical applications. Secondly, we explore the impact of whether to distinguish directions when generating topology for diagram classification. Compared with the generation method introduced in \ref{generation of the logic structure}, we simply use lines of the same thickness between circles to indicate all relations between objects. TABLE \ref{acc per cato} shows the classification accuracy of each category when only directed or undirected topology is used. The results show that the performance of directed topology is better than undirected topology in general. The improvements are particularly obvious in some categories where the relations are directional. For example, the accuracy of the directed graph is increased by 18.18\%, and the accuracy of the linked list is increased by 11.11\%. In the remaining undirected categories, there is no obvious difference in performance between the two methods. It can be concluded that the use of a directed topology is advantageous for analyzing the relation information in the diagrams.

\section{Conclusion}
\label{conclusion}

We introduce a type of diagram only containing geometric shapes and construct the first dataset CSDia of this type with rich annotations. Based on this dataset, we study the representation and understanding of the diagram. We propose a novel method of generating the topology of the diagram to extract the topological structures and remove visual noises. We propose the DPN model, which analyzes diagram, topology and text in parallel, and use it on the classification task to test the ability of understanding the diagram. Our experimental results show improvements of DPN in understanding diagrams compared to other baselines. Moreover, We further study the role and impact of the topology branch in DPN.

We have released the CSDia dataset and the DPN model on github\footnote{https://github.com/WayneWong97/CSDia}, which provides convenience for other researchers to do further work on diagram object dectection, textbook question answering and so on. These studies will be conductive to the understanding of multimodal knowledge, so as to the development of the intelligent education.

\section*{Acknowledgment}
This work was supported by National Key Research and
Development Program of China (2018YFB1004500), National
Natural Science Foundation of China (61877050), Innovative Research
Group of the National Natural Science Foundation of China
(61721002), Innovation Research Team of Ministry of Education (IRT 17R86), Project of China Knowledge Centre for
Engineering Science and Technology, MoE-CMCC ”Artifical
Intelligence” Project (MCM20190701), The National Social
Science Fund of China(18XXW005), National Statistical Science Research Project(2020LY103), Ministry of Education
Humanities and Social Sciences Fund(17YJA860028).

\ifCLASSOPTIONcaptionsoff
  \newpage
\fi



%

\bibliographystyle{IEEEtran}
\bibliography{./reference}

\begin{thebibliography}{10}
\providecommand{\url}[1]{#1}
\csname url@samestyle\endcsname
\providecommand{\newblock}{\relax}
\providecommand{\bibinfo}[2]{#2}
\providecommand{\BIBentrySTDinterwordspacing}{\spaceskip=0pt\relax}
\providecommand{\BIBentryALTinterwordstretchfactor}{4}
\providecommand{\BIBentryALTinterwordspacing}{\spaceskip=\fontdimen2\font plus
\BIBentryALTinterwordstretchfactor\fontdimen3\font minus
  \fontdimen4\font\relax}
\providecommand{\BIBforeignlanguage}[2]{{%
\expandafter\ifx\csname l@#1\endcsname\relax
\typeout{** WARNING: IEEEtran.bst: No hyphenation pattern has been}%
\typeout{** loaded for the language `#1'. Using the pattern for}%
\typeout{** the default language instead.}%
\else
\language=\csname l@#1\endcsname
\fi
#2}}
\providecommand{\BIBdecl}{\relax}
\BIBdecl

\bibitem{lu2007survey}
D.~Lu and Q.~Weng, ``A survey of image classification methods and techniques
  for improving classification performance,'' \emph{International journal of
  Remote sensing}, vol.~28, no.~5, pp. 823--870, 2007.

\bibitem{long2015fully}
J.~Long, E.~Shelhamer, and T.~Darrell, ``Fully convolutional networks for
  semantic segmentation,'' in \emph{Proceedings of the IEEE conference on
  computer vision and pattern recognition}, 2015, pp. 3431--3440.

\bibitem{antol2015vqa}
S.~Antol, A.~Agrawal, J.~Lu, M.~Mitchell, D.~Batra, C.~Lawrence~Zitnick, and
  D.~Parikh, ``Vqa: Visual question answering,'' in \emph{Proceedings of the
  IEEE international conference on computer vision}, 2015, pp. 2425--2433.

\bibitem{lin2014microsoft}
T.-Y. Lin, M.~Maire, S.~Belongie, J.~Hays, P.~Perona, D.~Ramanan,
  P.~Doll{\'a}r, and C.~L. Zitnick, ``Microsoft coco: Common objects in
  context,'' in \emph{European conference on computer vision}, 2014, pp.
  740--755.

\bibitem{young2014image}
P.~Young, A.~Lai, M.~Hodosh, and J.~Hockenmaier, ``From image descriptions to
  visual denotations: New similarity metrics for semantic inference over event
  descriptions,'' \emph{Transactions of the Association for Computational
  Linguistics}, vol.~2, pp. 67--78, 2014.

\bibitem{krishna2017visual}
R.~Krishna, Y.~Zhu, O.~Groth, J.~Johnson, K.~Hata, J.~Kravitz, S.~Chen,
  Y.~Kalantidis, L.-J. Li, D.~A. Shamma \emph{et~al.}, ``Visual genome:
  Connecting language and vision using crowdsourced dense image annotations,''
  \emph{International journal of computer vision}, vol. 123, no.~1, pp. 32--73,
  2017.

\bibitem{atrey2010multimodal}
P.~K. Atrey, M.~A. Hossain, A.~El~Saddik, and M.~S. Kankanhalli, ``Multimodal
  fusion for multimedia analysis: A survey,'' \emph{Multimedia systems},
  vol.~16, no.~6, pp. 345--379, 2010.

\bibitem{kembhavi2017you}
A.~Kembhavi, M.~Seo, D.~Schwenk, J.~Choi, A.~Farhadi, and H.~Hajishirzi, ``Are
  you smarter than a sixth grader? textbook question answering for multimodal
  machine comprehension,'' in \emph{Proceedings of the IEEE Conference on
  Computer Vision and Pattern Recognition}, 2017, pp. 4999--5007.

\bibitem{watanabe1998diagram}
Y.~Watanabe and M.~Nagao, ``Diagram understanding using integration of layout
  information and textual information,'' in \emph{COLING 1998 Volume 2: The
  17th International Conference on Computational Linguistics}, 1998.

\bibitem{ferguson2000georep}
R.~W. Ferguson and K.~D. Forbus, ``Georep: A flexible tool for spatial
  representation of line drawings,'' in \emph{AAAI/IAAI}, 2000, pp. 510--516.

\bibitem{ferguson1998telling}
R.~Ferguson and K.~D. Forbus, ``Telling juxtapositions: Using repetition and
  alignable difference in diagram understanding,'' \emph{Advances in Analogy
  Research}, pp. 109--117, 1998.

\bibitem{futrelle2003extraction}
R.~P. Futrelle, M.~Shao, C.~Cieslik, and A.~E. Grimes, ``Extraction, layout
  analysis and classification of diagrams in pdf documents,'' in \emph{Seventh
  International Conference on Document Analysis and Recognition, 2003.
  Proceedings.}, 2003, pp. 1007--1013.

\bibitem{seo2014diagram}
M.~J. Seo, H.~Hajishirzi, A.~Farhadi, and O.~Etzioni, ``Diagram understanding
  in geometry questions,'' in \emph{Twenty-Eighth AAAI Conference on Artificial
  Intelligence}, 2014.

\bibitem{sachan2017learning}
M.~Sachan and E.~Xing, ``Learning to solve geometry problems from natural
  language demonstrations in textbooks,'' in \emph{Proceedings of the 6th Joint
  Conference on Lexical and Computational Semantics (* SEM 2017)}, 2017, pp.
  251--261.

\bibitem{kembhavi2016diagram}
A.~Kembhavi, M.~Salvato, E.~Kolve, M.~Seo, H.~Hajishirzi, and A.~Farhadi, ``A
  diagram is worth a dozen images,'' in \emph{European Conference on Computer
  Vision}, 2016, pp. 235--251.

\bibitem{gomez2019look}
J.~M. Gomez-Perez and R.~Ortega, ``Look, read and enrich-learning from
  scientific figures and their captions,'' in \emph{Proceedings of the 10th
  International Conference on Knowledge Capture}, 2019, pp. 101--108.

\bibitem{gomez2020isaaq}
P.~Gomez, M.~Jose, and R.~Ortega, ``Isaaq--mastering textbook questions with
  pre-trained transformers and bottom-up and top-down attention,'' in
  \emph{Conference on Empirical Methods in Natural Language Processing}, 2020.

\bibitem{kim2018textbook}
D.~Kim, S.~Kim, and N.~Kwak, ``Textbook question answering with multi-modal
  context graph understanding and self-supervised open-set comprehension,'' in
  \emph{Proceedings of the Association for Computational Linguistics}, 2019.

\bibitem{morris2020slideimages}
D.~Morris, E.~M{\"u}ller-Budack, and R.~Ewerth, ``Slideimages: A dataset for
  educational image classification,'' in \emph{European Conference on
  Information Retrieval}, 2020, pp. 289--296.

\bibitem{sandler2018mobilenetv2}
M.~Sandler, A.~Howard, M.~Zhu, A.~Zhmoginov, and L.-C. Chen, ``Mobilenetv2:
  Inverted residuals and linear bottlenecks,'' in \emph{Proceedings of the IEEE
  conference on computer vision and pattern recognition}, 2018, pp. 4510--4520.

\bibitem{shaffer2012data}
C.~A. Shaffer, ``Data structures and algorithm analysis,'' \emph{Update},
  vol.~3, pp. 0--3, 2012.

\bibitem{mehlhorn2008algorithms}
K.~Mehlhorn and P.~Sanders, \emph{Algorithms and Data Structures: The Basic
  Toolbox}.\hskip 1em plus 0.5em minus 0.4em\relax Springer Science \& Business
  Media, 2008.

\bibitem{shuaihui2018}
H.~Shuai, \emph{High Score Notes of Data Structure}.\hskip 1em plus 0.5em minus
  0.4em\relax China Machine Press, 2018.

\bibitem{yanweimin2002}
W.~Yan and M.~Wu, \emph{Data Structure C version}.\hskip 1em plus 0.5em minus
  0.4em\relax TsingHua University Press, 2002.

\bibitem{tangxiaodan2007caozuoxitong}
X.~Tang, H.~Liang, F.~Zhe, and Z.~Tang, \emph{Computer Operating System}.\hskip
  1em plus 0.5em minus 0.4em\relax Xidian University Press, 2007.

\bibitem{tangshuofei2000zucheng}
S.~Tang, X.~Liu, and C.~Wang, \emph{Principles of Computer Organization}.\hskip
  1em plus 0.5em minus 0.4em\relax Higher Education Press, 2000.

\bibitem{liuchangshu2002shuzi}
C.~Liu, \emph{Digital Logic Circuit}.\hskip 1em plus 0.5em minus 0.4em\relax
  National Defense Industry Press, 2002.

\bibitem{simonyan2014very}
K.~Simonyan and A.~Zisserman, ``Very deep convolutional networks for
  large-scale image recognition,'' in \emph{Proceedings of the IEEE conference
  on Computer Vision and Pattern Recognition}, 2014.

\bibitem{pennington2014glove}
J.~Pennington, R.~Socher, and C.~D. Manning, ``Glove: Global vectors for word
  representation,'' in \emph{Proceedings of the 2014 conference on empirical
  methods in natural language processing (EMNLP)}, 2014, pp. 1532--1543.

\bibitem{deng2009imagenet}
J.~Deng, W.~Dong, R.~Socher, L.-J. Li, K.~Li, and L.~Fei-Fei, ``Imagenet: A
  large-scale hierarchical image database,'' in \emph{2009 IEEE conference on
  computer vision and pattern recognition}, 2009, pp. 248--255.

\bibitem{easyocr}
J.~AI, ``Easyocr,'' \url{https://github.com/JaidedAI/EasyOCR}, 2020.

\bibitem{he2016deep}
K.~He, X.~Zhang, S.~Ren, and J.~Sun, ``Deep residual learning for image
  recognition,'' in \emph{Proceedings of the IEEE conference on computer vision
  and pattern recognition}, 2016, pp. 770--778.

\bibitem{xie2017aggregated}
S.~Xie, R.~Girshick, P.~Doll{\'a}r, Z.~Tu, and K.~He, ``Aggregated residual
  transformations for deep neural networks,'' in \emph{Proceedings of the IEEE
  conference on computer vision and pattern recognition}, 2017, pp. 1492--1500.

\bibitem{iandola2016squeezenet}
F.~N. Iandola, S.~Han, M.~W. Moskewicz, K.~Ashraf, W.~J. Dally, and K.~Keutzer,
  ``Squeezenet: Alexnet-level accuracy with 50x fewer parameters and $<$0.5 mb
  model size,'' \emph{Proceedings of the IEEE Conference on Computer Vision and
  Pattern Recognition}, 2017.

\end{thebibliography}

\end{document}